\documentclass[sn-basic]{sn-jnl}
\usepackage{caption}
\usepackage{graphicx}%
\usepackage{multirow}%
\usepackage{amsmath,amssymb,amsfonts}%
\usepackage{amsthm}%
\usepackage{mathrsfs}%
\usepackage[title]{appendix}%
\usepackage{xcolor}%
\usepackage{textcomp}%
\usepackage{manyfoot}%
\usepackage{booktabs}%
\usepackage{algorithm}%
\usepackage{algorithmicx}%
\usepackage{algpseudocode}%
\usepackage{listings}%
\usepackage{tikz}
\usepackage{hyperref} 

\begin{document}

\title[Article Title]{Predicting Male Fertility Using Machine Learning: A Semen Parameters Based Analysis with the VISEM Dataset}
\author[1]{\fnm{Shahnawaz} \sur{Qureshi}
}
\email{shahnawaz.qureshi@paf-iast.edu.pk}
\author*[2]{\fnm{Raja Khurram} \sur{Shahzad} 
}
\email{raja-khurram.shahzad@miun.se}
\author[1]{\fnm{Muhammad} \sur{Fozan}} \email{mfozan67@gmail.com}
\author[1]{\fnm{Emal} \sur{Kawal}}\email{b22f1813cs118@fecid.paf-iast.edu.pk}
\author[3]{\fnm{Syed} \sur{Aziz Shah}
}
\email{syed.shah@coventry.ac.uk}
\author[4]{\fnm{Sattam} \sur{Al-Anazi}}
\email{s.homidi@saso.gov.sa}
\author[5]{\fnm{Syed Muhammad} \sur{Zeeshan Iqbal} 
}
\email{mzeeshan01@gmail.com}

\affil[1]{\orgdiv{School of Computing}, \orgname{Pak-Austria Fachhochschule: Institute of Applied Sciences and Technology}, \orgaddress{\city{Haripur}, \state{Khyber Pakhtunkhwa}, \country{Pakistan}}}
\affil[2]{\orgdiv{Department of Communication, Quality Management and Information Systems}, \orgname{Mid Sweden University}, \orgaddress{\city{Östersund Campus}}, \country{Sweden}}
\affil[3]{\orgname{Healthcare Sensing Technology, Center for Intelligent Healthcare, Coventry University}, \orgaddress{\country{United Kingdom}}}
\affil[4]{\orgname{E-Serivces Department, Saudi Standards, Metrology and Quality Organization}, \orgaddress{\city{Riyadh}, \country{Saudi Arabia}}}
\affil[5]{\orgname{Research and Development, Brightware LLC}, \orgaddress{\city{Riyadh}, \country{Saudi Arabia}}}

\abstract{
Male infertility is a significant yet often underdiagnosed aspect of reproductive health, with semen analysis serving as the cornerstone of clinical evaluation. To address this problem, this study investigates the use of machine learning algorithms to classify male fertility status based on key semen parameters, i.e.,  sperm concentration, motility, and morphology, using the VISEM dataset. This dataset includes semen samples from 85 participants, classified into three categories, i.e., Fertile, Sub-Fertile, and Infertile, according to the World Health Organization's criteria. After pre-processing and feature engineering, the dataset was used to train and assess multiple classification models using the LazyPredict framework. Among the more than 40 algorithms tested, the Nearest Centroid classifier achieved an accuracy of 94.2\%, outperforming other models such as Support Vector Machines and Quadratic Discriminant Analysis. The model's robustness was validated using 5-fold cross-validation and multiclass ROC-AUC analysis. This study illustrates that machine learning models can provide fast, accurate, and objective assessments of semen quality, potentially supporting clinical decision-making in andrology and assisted reproductive technologies. These findings emphasize the growing potential of machine learning to enhance fertility diagnostics and inform patient-specific treatment strategies.
}
\keywords{Male infertility, Semen analysis, Sperm motility, Sperm morphology, Machine learning, Fertility classification, VISEM dataset, Artificial intelligence in reproduction, Reproductive health, Clinical decision support, Reproductive diagnostics}
\maketitle
\section{Introduction}\label{Intro}
\begin{figure*}[!htbp]
\centering
\includegraphics[width=0.6\textwidth]{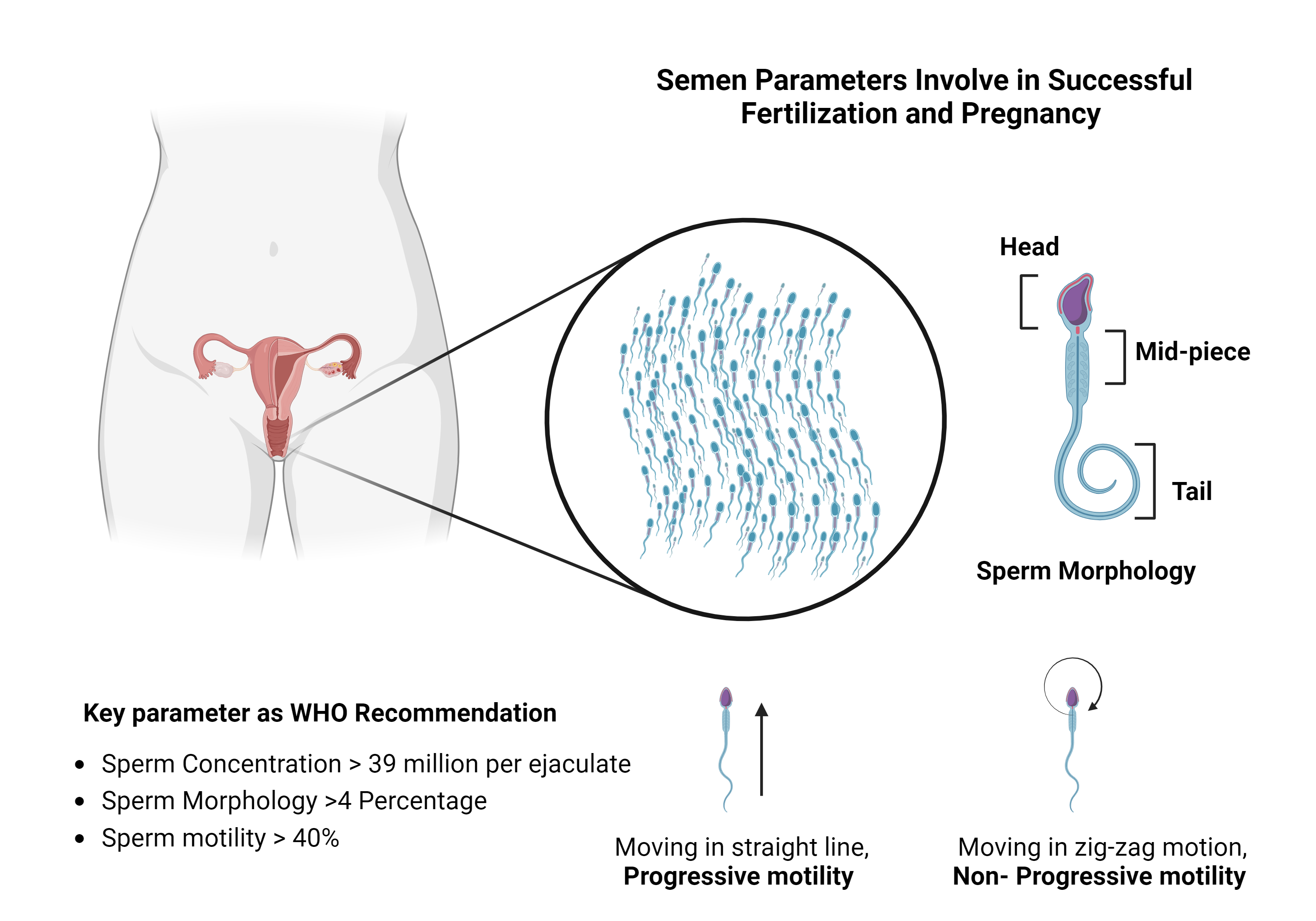}
\caption{This figure presents essential semen parameters for successful fertilization according to WHO recommendations. Key thresholds include normal sperm morphology above 4\%, sperm concentration above 39 million per ejaculation, and total motility above 40\%. It distinguishes between progressive motility (forward movement) and non-progressive motility (erratic motion, zig-zag motion). The diagram also highlights the main components of sperm,  i.e., head, midpiece, and tail, emphasizing their roles in reaching and penetrating the oocyte during fertilization.}
\label{fig:Pict1}
\end{figure*}

Infertility is defined as the inability to conceive after 12 months of unprotected intercourse and affects approximately 17.5\% of the global adult population. Male factors are responsible for nearly half of all infertility cases, occurring either independently (up to 20\%) or in combination with female infertility (up to 40\%) \cite{vander2018fertility}. Semen analysis is the primary clinical tool for assessing male reproductive health. It is used to evaluate critical parameters, such as sperm concentration, motility, and morphology, according to guidelines established by the World Health Organization (WHO) \cite{kumar2015trends}, as illustrated in the Figure \ref{fig:Pict1} . However, despite its clinical significance, traditional semen evaluation is performed manually and is subject to observer bias, inter-laboratory variability, and inconsistent interpretation. These factors may result in inaccurate or delayed diagnoses  \cite{leslie2024male}.

As fertility evaluations become increasingly complex, artificial intelligence (AI) and machine learning (ML) are emerging as transformative tools in reproductive medicine. These technologies enable objective analysis of diverse datasets and have demonstrated strong predictive capabilities for classifying male fertility potential. While previous studies have focused on predicting motility or performing image-based analyses, there has been a limited systematic evaluation of the effectiveness of ML models in categorizing fertility status into clinically relevant groups using standardized semen attributes \cite{ottl2022motilitai, Nguyen2023, kobayashi2024new, Tiab2023}.

This study addresses that gap by applying supervised ML models to the publicly available VISEM dataset\footnote{https://datasets.simula.no/visem/}, which contains semen analysis records from 85 male participants. The primary aim is to classify these samples into three fertility categories, i.e., Fertile, Sub-Fertile, and Infertile, based on WHO thresholds \cite{world1999laboratory}. Using the LazyPredict framework\footnote{https://pypi.org/project/lazypredict}, we evaluated over 40 ML algorithms to identify the most accurate model, assessing performance through cross-validation\footnote{https://en.wikipedia.org/wiki/Cross-validation\_(statistics)}
and Receiver Operating Characteristic - Area Under Curve (ROC-AUC)  analysis\footnote{https://en.wikipedia.org/wiki/Receiver\_operating\_characteristic}. Our findings suggest that ML techniques, particularly the Nearest Centroid classifier\footnote{https://en.wikipedia.org/wiki/Nearest\_centroid\_classifier}, can significantly enhance the accuracy and consistency of fertility assessments. This research has important implications for andrology and assisted reproductive technologies (ART), providing a robust decision-support tool to assist clinicians and couples in fertility planning.

The remainder of this paper is organized as follows: Section \ref{sec:background} provides an overview of human fertilization and semen evaluation. Section \ref{sec:related_work} reviews previous AI/ML-based approaches. Section \ref{sec:methodology} presents the dataset and methods used in the study. Section \ref{sec:results} presents the experimental results. Section \ref{sec:discussion} discusses the findings, and Section \ref{sec:conclusion} concludes the paper with conclusions and future directions.
\section{Background}\label{sec:background}
Male reproductive health is influenced by a range of physiological, genetic, and environmental factors, as illustrated in the Figure \ref{fig:Pict1}. The human spermatozoon, or sperm, is a specialized motile cell that must travel an average distance of approximately 19 cm in the female reproductive tract to fertilize the oocyte \cite{dcunha2022current}. Sperm are generally categorized into progressively motile, non-progressive, and immotile types, with progressive motility considered the strongest predictor of male fertility \cite{brown1944rate, world2021laboratory}. If the total motility is below 40\% or if progressive motility is under 32\%, both rates are strongly associated with reduced fertility, a condition known as asthenozoospermia \cite{Cavarocchi2025-zl}. \cite{cooper2010world, nayak2019oxidative, mclachlan2013approach, kao2004sperm}. In addition to motility, sperm morphology plays a key role in fertility potential. Abnormal morphology may be linked to genetic defects, such as aneuploidy or mitochondrial dysfunction, and structural deformities in the head or tail \cite{Darand2023, carlsen1992evidence}. According to WHO criteria, a morphology threshold of 4\% normal forms is used to define fertile samples \cite{world1999laboratory, zhou2021influence}. Sperm concentration is another vital parameter, with a normal range of 15 million sperm per mL or 39 million per ejaculate \cite{bonde1998relation}. A strong correlation exists between sperm concentration and the likelihood of conception; however, both partners' ages influence this relationship, particularly the female partner's, due to the declining quality of oocytes after age 35  \cite{bonde1998relation}. Advanced fertility treatments, such as in vitro fertilization (IVF),  require an evaluation of reproductive parameters for both partners. The success of IVF depends on accurate evaluation of these parameters, yet manual interpretations can introduce variability between clinics \cite{grow1994sperm, leslie2024male}. These challenges highlight the need for consistent, data-driven tools to support clinical decision-making in reproductive health.

\section{Related Work}\label{sec:related_work}
\subsection*{Challenges in Semen Analysis}
Male infertility remains a significant challenge in reproductive medicine, particularly due to the subjective nature of traditional semen analysis. These evaluations rely on manual microscopy and expert interpretation by andrologists, making the process labor-intensive and vulnerable to inter-observer variability \cite{Gbagbo2024}. This subjectivity leads to inconsistent diagnoses, especially across clinics with varying expertise and standardization. To address these limitations, artificial intelligence and machine learning have emerged as promising tools for enhancing diagnostic accuracy and consistency. In particular, the VISEM dataset, which contains annotated videos, images, and semen parameters, has become a benchmark resource for developing intelligent models in reproductive diagnostics.
\subsection*{Machine Learning for Fertility Prediction}
Most ML-based studies using the VISEM dataset have focused on predicting sperm motility as a continuous variable. For example, Nguyen et al. \cite{Nguyen2023} developed a deep learning model for estimating motility, achieving a Mean Absolute Error (MAE) of 9.322. They emphasized the importance of customized loss functions for clinical relevance. Similarly, Ottl et al. \cite{ottl2022motilitai} introduced \textit{motilitAI}, a hybrid framework combining support vector regression (SVR) and neural networks, achieving an MAE of 7.31 using engineered movement features. 

Particularly, Adinugroho et al. \cite{Adinugroho2023} introduced MotionFlow, a model that combines motion encoding and morphological features. This model achieved an MAE of 6.842\% for motility and 4.148\% for morphology. However, the focus of these efforts has primarily been on regressing individual parameters rather than on the comprehensive classification of fertility status. Specifically, there is a lack of frameworks that categorize individuals as Fertile, Sub-Fertile, or Infertile based on WHO clinical thresholds. This indicates a gap in the literature regarding interpretable, multi-parameter fertility prediction frameworks.

\subsection*{Advances in Detection and Tracking}
Accurate detection and tracking are fundamental for Computer-Assisted Sperm Analysis (CASA) systems. Choi et al. \cite{Choi2022} demonstrated that CASA systems rely heavily on robust segmentation algorithms. Thambawita et al. \cite{Thambawita2023} used the VISEM-Tracking Robust dataset,  which consists of over 29,000 annotated frames, to facilitate the training of real-time detection systems using YOLOv5\footnote{https://github.com/ultralytics/yolov5}. Similarly, Dobrovolny et al. \cite{Dobrovolny2023} applied YOLOv5 to the VISEM dataset, achieving a mean average precision of 72.15\%. Valiuškaitė et al. \cite{Valiuskaitė2023} developed a  Rregion-based Convolutional Neural Network (R-CNN) that achieved a detection accuracy of 91.77\%, showing a strong correlation with sperm vitality (r = 0.969). Saadat et al. \cite{Saadat2023} further demonstrated that UNet++ with ResNet34 performs well in segmenting sperm heads, although challenges remain in distinguishing sperm from adjacent non-sperm cells.
\subsection*{Feature Engineering and Model Architectures}
Several studies have highlighted the importance of engineered features and advanced model architectures for improving sperm classification. Ottl et al. \cite{Ottl2021} compared the performance of Support Vector Regression, Convolutional Neural Networks (CNN), and Recurrent Neural Networks (RNN) for tracking sperm motion using feature aggregation techniques such as Crocker-Grier vectors\cite{CROCKER_1996} and bag-of-words methods. More recently, Tiab et al. \cite{Tiab2023} evaluated MobileNet, YOLOv5s, and DeepSort across the VISEM and another dataset, achieving a classification accuracy of 99.2\% and a Multi-Object Tracking Precision of 99\%. This research confirms the effectiveness of transfer learning approaches and the benefits of dataset diversity.
\subsection*{Gaps in Literature and Our Contribution}
While the use of artificial intelligence (AI) and machine learning in andrology is increasing, many existing studies often focus on predicting isolated parameters such as motility or morphology, rather than clinically meaningful fertility categorization. Moreover, most of these models optimize for predictive accuracy without accounting for interpretability or integration with clinical decision-making workflows. These limitations restrict their real-world applicability, particularly in assisted reproductive technologies (ART) and fertility counseling.

In this study, we address these existing gaps by implementing and comparing over 40 supervised ML classifiers using the LazyPredict framework \cite{musthyala2024ai} on the VISEM dataset. We classify samples into Fertile, Sub-Fertile, and Infertile groups based on WHO thresholds for concentration, morphology, and progressive motility. Our evaluation includes metrics such as precision, recall, F1-score, and AUC to identify robust, interpretable models that can support clinical decision-making. This work contributes to the literature by shifting the focus from isolated parameter predictions to a comprehensive classification of fertility-relevant parameters for clinical settings.
\begin{figure*}[!htbp]
\centering
\includegraphics[width=0.8\textwidth]{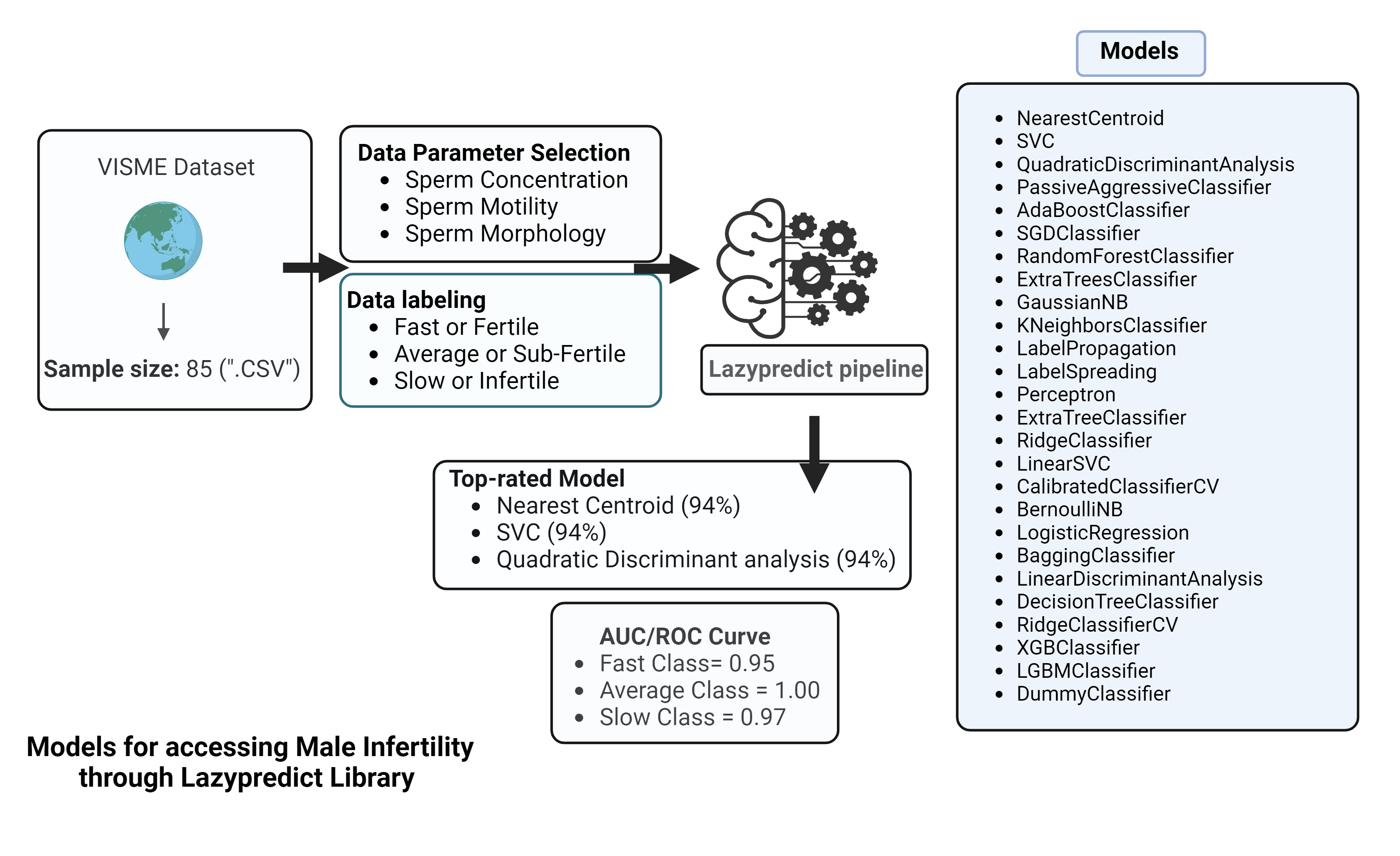}
\caption{\textbf{Workflow for Fertility Classification Using Semen Parameters and Machine Learning}. This flowchart presents a method for classifying male fertility status using semen parameters. The LazyPredict framework evaluated and ranked 40 machine learning models based on cross-validated accuracy on the VISEM dataset.}
\label{fig:Pict2}
\end{figure*}
\section{Methodology}\label{sec:methodology}
This study used the VISEM dataset, comprising 85 semen samples, to develop machine learning models for classifying fertility based on parameters defined by the World Health Organization: sperm concentration, morphology, and progressive motility. After preprocessing the data and labeling the samples, we benchmarked 40 classifiers using LazyPredict. The top-performing models were Nearest Centroid, Support Vector Machine (SVM), and Quadratic Discriminant Analysis (QDA), which achieved the highest accuracies. To evaluate model performance, we employed 5-fold cross-validation and performed ROC-AUC analysis. Moreover, correlation studies revealed strong interdependencies among the seminal parameters that are important for predicting fertility. The overall workflow is illustrated in Figure~\ref{fig:Pict2}.

\subsection*{Dataset Description}
The VISEM dataset \cite{haugen2019visem} was developed by the Simula Research Laboratory in Norway. The dataset was created to facilitate research on the relationship between semen quality and male fertility. It contains semen analysis data from 85 healthy male participants aged 18 or older. All samples were evaluated by trained embryologists using a 40x magnification microscope with a heated stage maintained at 37\textdegree C to simulate physiological conditions. The dataset comprises various quantitative semen parameters, including sperm concentration, total sperm count, sperm motility, sperm morphology, ejaculate volume, and sperm viability \cite{haugen2019visem,thambawita2021deep}. 

\subsection*{Data Preprocessing and Feature Engineering}
As shown in Figure~\ref{fig:Pict2}, preprocessing was performed to remove unnecessary spaces and missing values, ensuring consistency across all samples. A unique  identifier (ID) was assigned to each entry to maintain data integrity. The following key features were extracted, normalized, and engineered for training purposes:
\begin{itemize}
     \item \textbf{ID}: A unique identifier assigned to each semen sample.
    \item \textbf{Sperm concentration}: The number of sperm per milliliter, which reflects testicular function.
    \item \textbf{Total sperm count}: The total number of sperm in the ejaculate, calculated by multiplying sperm concentration by ejaculate volume.
    \item \textbf{Normal spermatozoa}: The percentage of morphologically normal sperm, determined using the \textbf{Teratozoospermia Index (TZI)}:
 \[
  \text{TZI} = \frac{h + m + t}{ab}
  \]
  where \( h \) = head defects, \( m \) = mid-piece defects, \( t \) = tail defects, and \( ab \) = total abnormal sperm \cite{carlsen1992evidence}.
\item \textbf{Progressive motility}: The percentage of sperm that are moving actively in a straight line or in large circles \cite{world2021laboratory}.
    \item \textbf{Non-progressive motility}: The percentage of sperm exhibiting erratic or low-energy movement.
    \item \textbf{Immotile sperm}: The percentage of sperm showing no movement, which indicates low vitality.

\end{itemize}

\subsection*{Fertility Classification Criteria}
To categorize male fertility potential, we used classification thresholds based on the WHO's 6\textsuperscript{th} edition manual for semen evaluation \cite{world2021laboratory}. The dataset was labeled into three fertility classes, i.e, Fertile, Sub-Fertile, and Infertile, based on progressive motility ($\geq 40\%$, $20$--$39\%$, and $< 20\%$), morphology ($\geq 4\%$, $2$--$3\%$, and $<2\%$), and sperm concentration ($\geq 15$, $10$--$14$, and $<10$ million/mL) \cite{cooper2010world}, as illustrated in the table \ref{tab:fertility}. Non-progressive motility was excluded, as studies suggest it contributes minimally to live birth outcomes \cite{zhou2021influence}.
\begin{table}[htbp]
    \centering
      \caption{Semen Characteristics and Fertility Classifications for Model Development.}
    \begin{tabular}
    {|p{3cm}|p{2cm}|p{1.95cm}|p{3.25cm}|}
    \hline
    \textbf{Condition} & \textbf{Progressive Motility (\%)} & \textbf{Normal Morphology (\%)} & \textbf{Sperm Concentration (million/mL)} \\
    \hline
    Fertile or Fast       & $\geq 40\%$     & $\geq 4\%$     & $\geq 15$ million/mL  \\
    \hline
    Sub-Fertile or Average & $>20\%$ and $<39\%$ & 2\%-3\%        & 10-14 million/mL      \\
    \hline
    Infertile or Slow     & $<20\%$         & 2\%            & $<10$ million/mL      \\
    \hline
    \end{tabular}

    \label{tab:fertility}
\end{table}

\subsection*{Machine Learning Models' Evaluation}
The performance of the models was evaluated using accuracy, F1-score, and balanced accuracy, all calculated through 5-fold cross-validation \cite{raschka2018model}. To test for statistical significance, we performed a paired t-test with a significance threshold of \(p < 0.05\). To address sample dependence across the folds, a correction factor was utilized \cite{dietterich1998approximate}. Models that achieved over 90\% accuracy demonstrated a significant improvement compared to the ZeroR baseline, which is a na\"{\i}ve classifier that predicts only the majority class.

\subsection*{LazyPredict Benchmarking}

For our machine learning baseline, we used the LazyPredict  \cite{musthyala2024ai} package to evaluate the performance of several well-known machine learning algorithms on the semen analysis data. LazyPredict automates the training and testing of over 40 models and ranks them according to performance metrics such as accuracy, F1-score, and balanced accuracy. 
The models that achieved the highest performance in this experiment are Nearest Centroid, Support Vector Machine,  Quadratic Discriminant Analysis,  Gaussian Naive Bayes, and Random Forest. LazyPredict streamlines the modeling process by automatically feeding data into various models without the need for hyperparameter tuning. It offers performance metrics for easy comparison. Model performance is evaluated using 5-fold cross-validation, and models are ranked according to their average cross-validation accuracy. The framework uses a simple calculation to effectively evaluate and compare the performance of each model, as illustrated in Equation \ref{eq:PS} :

\begin{equation} \label{eq:PS}
    \text{Performance Score} = \frac{\text{True Positives} + \text{True Negatives}}{\text{Total Sample}}
\end{equation}

\subsubsection* {Nearest Centroid Model}
The Nearest Centroid classifier assigns samples to a class based on their distance to the centroid of each class in the feature space. The centroid is calculated by averaging the feature vectors of all samples within that class, as shown in Equation \ref{eq:NCM} \cite{garcia2018regression}.
\begin{equation} \label{eq:NCM}
    \text{Centroid} = \frac{1}{n} \sum_{i=1}^{n} X_i
\end{equation}
Where \( X_i \) refers to the feature vectors of each sample, while \( n \) denotes the total number of samples in the class. 
\subsubsection* {Support Vector Machine}
A Support Vector Machine creates a hyperplane in a high-dimensional space to effectively separate different classes. It maximizes the margin between these classes by solving the related optimization problem, as shown in Equation \ref{eq:SVM}:
\begin{equation} \label{eq:SVM}
    \min \left( \frac{1}{2} \|w\|^2 + C \sum_{i=1}^{n} \max(0, 1 - y_i(w \cdot X_i + b)) \right)
\end{equation}
In this context, \( w \) represents the weight vector, \( C \) is the regularization parameter, and \( b \) represents the bias. Support Vector Machines are particularly effective when samples are expected to be closely clustered in the feature space \cite{cortes1995support}. This concept is particularly relevant in studies of sperm morphology and motility, where scores can vary significantly.  
\subsubsection* {Quadratic Discriminant Analysis}
Quadratic Discriminant Analysis (QDA) was used to address the classification problem, assuming that different classes follow a normal distribution and that the decision boundary between them is quadratic in shape \cite{ghojogh2019linear}. The posterior probability for each class is calculated using Bayes' theorem as shown in Equation \ref{eq:QDA}: \newline
\begin{equation}\label{eq:QDA}
    P(y = k \mid x) = \frac{P(x \mid y = k) P(y = k)}{\sum_{l} P(x \mid y = l) P(y = l)}
\end{equation}
This probabilistic model assumes features are normally distributed and allows for non-linear class boundaries \cite{ghojogh2019linear}. Thus, QDA is especially useful for assessing sperm parameters, as these parameters are assumed to follow a normal distribution. 
\subsubsection* {Gaussian Naive Bayes}
The Gaussian Naive Bayes (GaussianNB) classifier uses Bayes’ Theorem to estimate the lthe probability of a class based on the measured features. GaussianNB assumes that the input predictors, such as sperm concentration, morphology, and motility, follow a normal distribution. Consequently, the model makes predictions on test samples based on this assumption. The probability is calculated using the normal probability density function, as shown in Equation \ref{eq:GNB}.
\begin{equation}\label{eq:GNB}
    P(X_i \mid C) = \frac{1}{\sqrt{2\pi\sigma^2}} \cdot e^{-\frac{(X_i - \mu)^2}{2\sigma^2}}
\end{equation}
Here, \(X_i\) represents the actual measured feature (for example, sperm concentration in millions per milliliter), \(\mu\)  is the mean of the features for class \(C\), and \(\sigma\) is the variance of the distribution of those features for that class. The GaussianNB classifier is known for its robustness, interpretability, and efficiency \cite{zhang2004optimality}.
\subsubsection* {Random forest Classifier}
The Random Forest Classifier is an adaptive learning technique that employs multiple decision trees to minimize the chances of errors in decision-making \cite{breiman2001random}. This method is particularly useful in systems where various factors must be considered to reach a conclusion, such as in semen analysis. In a Random Forest, the final prediction is made by averaging the outputs from all the decision trees in the forest, as shown in Equation \ref{eq:RF}.
\begin{equation}\label{eq:RF}
    \hat{y} = \arg\max_C \left( \sum_{j=1}^{m} T_j(X) \right)
\end{equation}
The term $T_j(X)$ refers to the prediction made by the j\textsuperscript{th} tree in the forest, and $m$ represents the total number of trees. 
\subsection* {Cross-Validation and Statistical Testing}
To minimize overfitting and evaluate model performance on unseen data, all models were assessed using 5-fold cross-validation. This technique is beneficial because it exposes models to various subsets of data, helping prevent overfitting. Moreover, the Naive Bayes model was hyperparameter-tuned using \texttt{GridSearchCV} to identify the optimal parameters. The primary focus during this tuning process was the \( \texttt{Var\_smoothing} \) parameter, which significantly enhanced the model's performance on the semen analysis dataset.
\subsection* { ROC-AUC Analysis }
In addition to basic performance measures that focus on classification accuracy, Receiver Operating Characteristic curves (ROC) were used to evaluate the model's performance. Since the dataset involves a multi-class problem, a one-vs-rest approach was used, treating each class as positive and the others as negative. The performance of the level one models was satisfactory, with no misclassifications. Consequently, the Area Under the Curve (AUC) was calculated for each class to assess the model's ability to differentiate between classes. 
The resulting Area Under the Curve scores were:
\begin{itemize}
\item Fast Class: AUC = 0.95
\item Average Class: AUC = 1.00
\item Slow Class: AUC = 0.97
\end{itemize}
The resulting AUC scores indicate that the classifier effectively distinguishes between different sperm motility classes.
\subsubsection* {Feature Visualization }
To investigate the relationships among progressive motility, morphology, and sperm concentration, we generated density plots for each feature. These plots helped us evaluate how effectively we could differentiate between various fertility classes and seminal characteristics. They supported the classification boundaries we selected \cite{zhou2021influence}, providing deeper insights into the factors that influence classification.

\section{Results}\label{sec:results}
This study examined semen samples from the VISEM dataset, which included 85 male participants aged 18 years and older. After preprocessing and labeling according to WHO standards, various machine learning classifiers were applied to predict fertility status.

\subsubsection* {Dataset Statistics and Class Distribution}
\begin{figure}[htbp]
\centering
\includegraphics[width=0.5\textwidth]{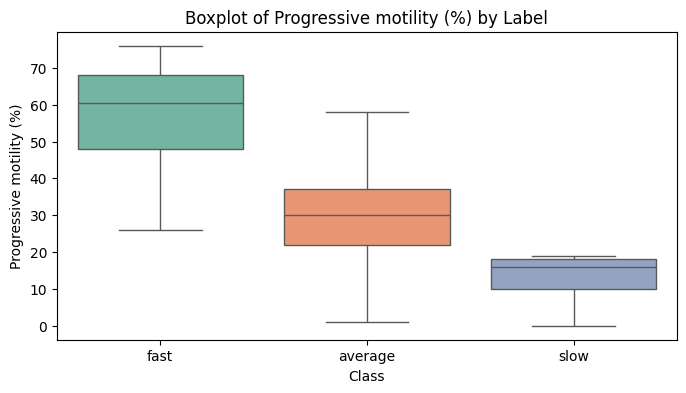}
\caption{\textbf{Progressive Motility Distribution Across Fertility Classes}.
This boxplot depicts the distribution of progressive motility percentages across three fertility classes: Fast, Average, and Slow. The Fast group exhibits the highest median progressive motility at 60\%, ranging from 4\% to 70\%. The Average group shows a median of 30\% with a broader range (0–50\%), while the Slow group has the lowest median at approximately 20\%, with most values between 10\% and 30\%. The visual representation clearly shows class separability, reinforcing the association between higher progressive motility and a superior fertility classification.
}
\label{fig:Pict3}
\end{figure}

The VISEM dataset comprises 85 samples and 21 attributes that capture essential sperm characteristics, including concentration, morphology, quality, motility, and deoxyribonucleic acid (DNA) fragmentation. Key columns in the dataset include sperm concentration (measured in $\times 10^{6}$ per ml), the percentage of normal spermatozoa, morphology, and progressive motility percentages. The label column categorizes each sample into one of three classes, i.e.,  Fast, Average, or Slow, based on these parameters as shown in Table  \ref{tab:class_distribution}.
\begin{table}[!htbp]
\centering
\caption{Fertility Class Distribution in the VISEM Dataset.}
\begin{tabular}{|l|c|}
\hline
\textbf{Class distribution} & \textbf{No.of Samples} \\
\hline
Fast or Fertile  & 46 \\
\hline
Average or Sub-Fertile  & 31 \\
\hline
Slow or Infertile & 08\\
\hline
\end{tabular}

\label{tab:class_distribution}
\end{table}
The dataset revealed that the sperm concentration (measured in millions per milliliter, $10^6/\text{ml}$) had a mean value of 
81.01 $\pm$ 65.34, with a range from 3 to 350. Progressive motility had a mean of 42.67\% $\pm$ 20.18\%, with a minimum of 0\% and a maximum of 76\%. Correlation analysis between parameters such as total sperm concentration, motility, and various morphological attributes indicated moderate to strong positive correlations. These correlations suggest multicollinearity, which can negatively affect model performance
(Please see the Figure~\ref{fig:Pict3}).

\begin{figure}[htbp]
\centering
\includegraphics[width=0.5\textwidth]{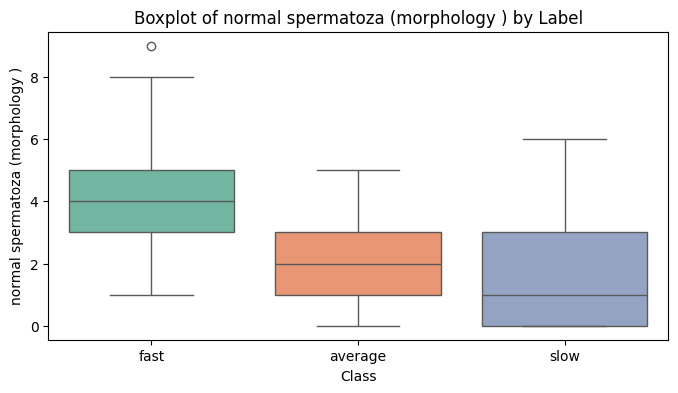}
\caption{\textbf{Normal Sperm Morphology Across Fertility Classes.} 
This boxplot displays the distribution of normal spermatozoa morphology percentages across three motility-based fertility classes: Fast, Average, and Slow. The Fast class has the highest median values, mostly ranging from 3\% to 6\%, with an outlier at 8.9\%. The Average class is centered between 1\% and 4\%, while the Slow class shows even lower percentages. This suggests a potential link between higher motility and improved sperm morphology, underscoring their combined importance in fertility assessment.
}
\label{fig:pict4}
\end{figure}

\subsubsection*{Feature Visualization and Interpretation}
\begin{figure}[htbp]
\centering
\includegraphics[width=0.5\textwidth]{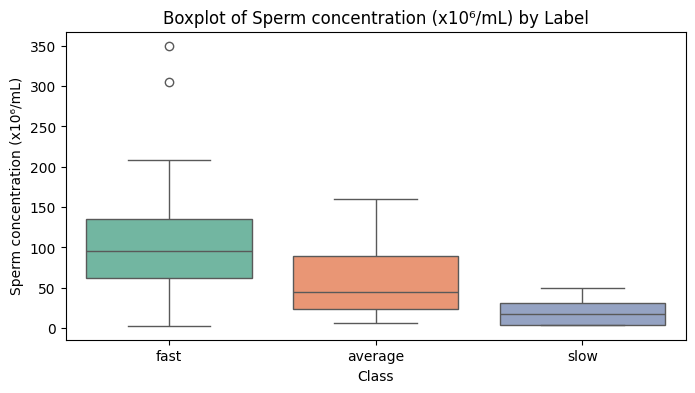}
\caption{\textbf{Sperm Concentration Across Fertility Classes}. This boxplot shows sperm concentration (millions per milliliter) across three fertility classes based on motility, i.e., Fast, Average, and Slow. The Fast group has a median concentration of about 50-150 million/mL, with some outliers exceeding 300 million/mL. The Average group ranges from 25 to 75 million/mL. The Slow group is around 25 million/mL with minimal variability. This visualization highlights the relationship between sperm concentration and motility-based fertility classification.}
\label{fig:Pict5}
\end{figure}

Figures~\ref{fig:Pict3}, \ref{fig:pict4}, and \ref{fig:Pict5} illustrate the distributions of progressive motility, morphology, and sperm concentration across fertility classes. The Fast class consistently showed higher median values in all three metrics. Specifically:

\begin{itemize}
    \item \textbf{Progressive Motility} (see Figure~\ref{fig:Pict3}): The Fast group had a median motility near 60\%, while the Average and Slow groups had median values of approximately 30\% and 20\%, respectively.
    \item \textbf{Sperm Morphology} (see Figure~\ref{fig:pict4}): The Fast group showed the highest median percentage of normal forms, with values reaching up to 8.9\%.
    \item \textbf{Sperm Concentration} (see Figure~\ref{fig:Pict5}): The Fast group had median concentrations ranging between 50 and 150 million/mL, which were significantly higher than those in the Average and Slow categories.
\end{itemize}

\noindent
These visualizations support the hypothesis that higher sperm quality, indicated by concentration, morphology, and motility, correlates with improved fertility classification.
\subsubsection*{Feature Correlation Analysis}
\begin{figure}[htbp]
\centering
\includegraphics[width=0.5\textwidth]{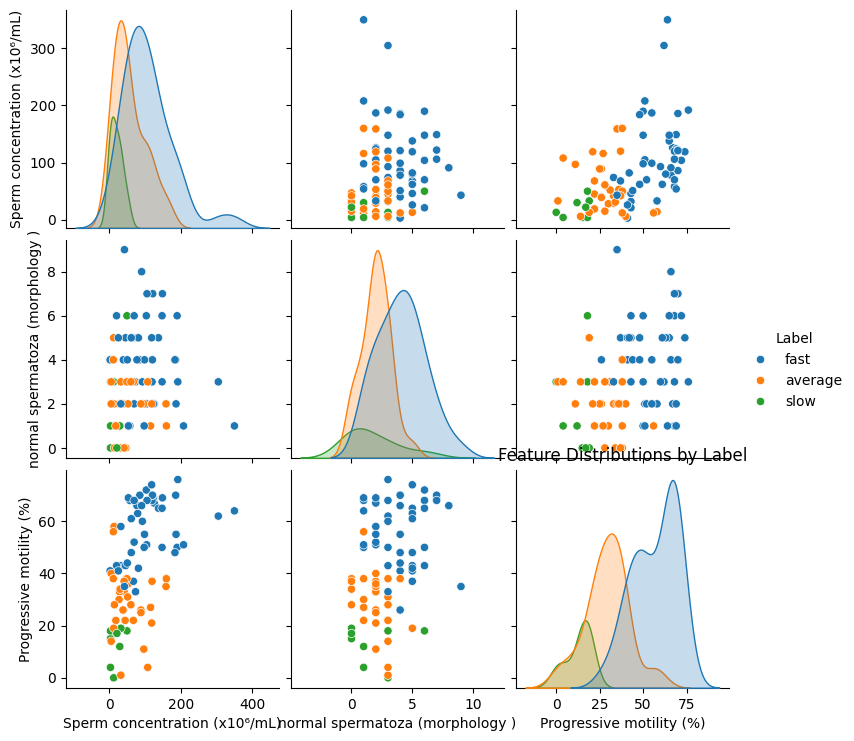}
\caption{%
\textbf{Correlation Between Sperm Parameters Across Fertility Classes.} 
This scatter plot matrix illustrates the relationships among sperm concentration (millions/mL), normal morphology (\%), and progressive motility (\%) across three fertility classes, i.e.,  Fast (blue), Average (orange), and Slow (green). The diagonal features kernel density plots that reveal distinct distributions for each parameter. Fast samples exhibit higher concentrations and motility, while Slow samples cluster at lower values, highlighting interdependencies among these fertility parameters.
}
\label{fig:Pict6}
\end{figure}

\begin{figure}[htbp]
\centering
\includegraphics[width=0.5\textwidth]{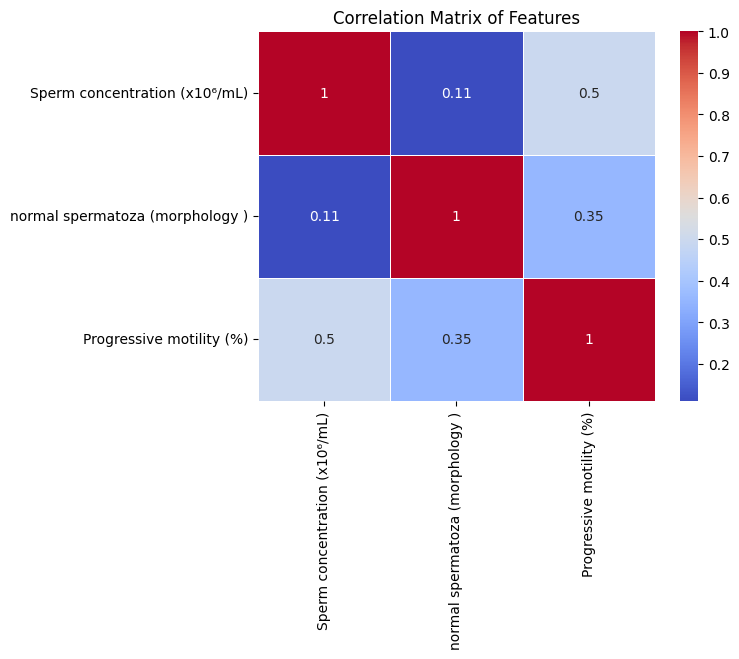}
\caption{%
\textbf{Correlation Heatmap of Key Sperm Parameters.} 
This heatmap shows the correlation matrix for sperm concentration (millions/mL), normal morphology (percentage), and progressive motility (percentage). There is a moderate positive correlation of 0.50 between concentration and motility, a weak positive correlation of 0.35 between morphology and motility, and a slight negative correlation of -0.11 between concentration and morphology. Color intensity indicates the strength of the correlation, from red (high positive) to blue (low or negative), illustrating the relationships among these factors in predicting fertility.}
\label{fig:Pict7}
\end{figure}

Pair plots (see Figure~\ref{fig:Pict6}) revealed clear separability among fertility classes based on sperm concentration, morphology, and motility. The diagonal kernel density plots illustrated distinct distributions: Fast samples (blue) exhibited higher values across all three dimensions, while Slow samples (green) were clustered at the lower end. Average samples (orange) fell in between. Figure~\ref{fig:Pict7} presents a heatmap illustrating the correlations among these three core features. Key findings include:
\begin{itemize}
    \item Sperm concentration and progressive motility: correlation = 0.50 (moderate positive)
    \item Morphology and motility: correlation = 0.35 (weaker positive)
    \item Concentration and morphology: correlation = –0.11 (slight negative)
\end{itemize}

\noindent
These results indicate that concentration and motility are more closely linked than morphology, although all three factors significantly contribute to fertility prediction.

\subsubsection*{Model Performance and Accuracy}
\begin{figure}[htbp]
\centering
\includegraphics[width=0.5\textwidth]{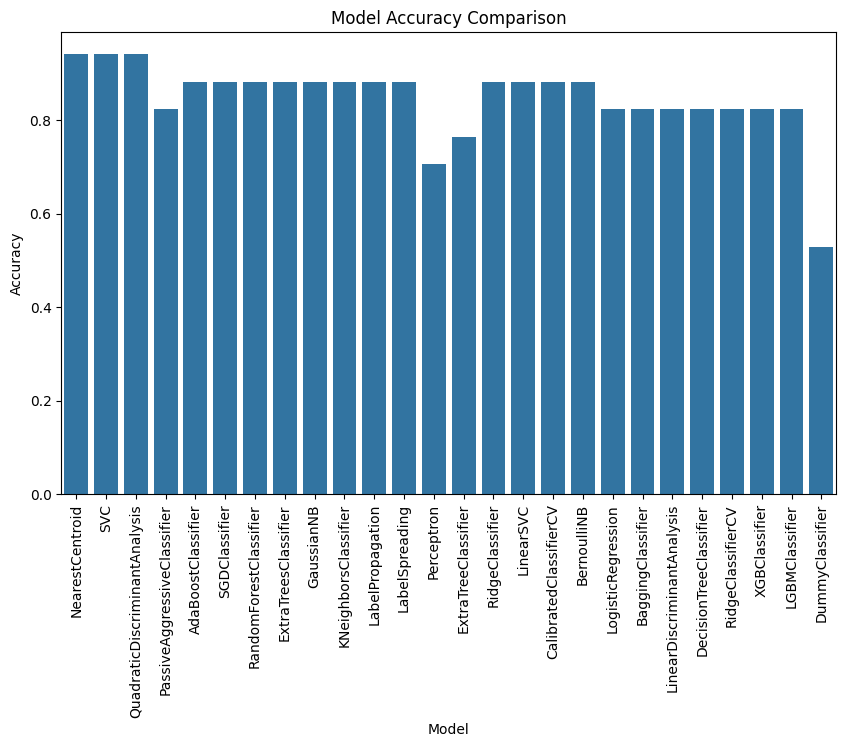}
\caption{%
\textbf{Accuracy of Machine Learning Models for Fertility Prediction.} 
This bar chart compares the performance of machine learning classifiers in predicting fertility status from seminal parameters. The top models ,i.e., Nearest Centroid, Support Vector Machine, and Quadratic Discriminant Analysis achieved over 90\% accuracy, while the ZeroR Classifier served as a lower baseline. Rankings are based on average cross-validation accuracy, demonstrating their effectiveness for real-world applications.}
\label{fig:Pict8}
\end{figure}

Using LazyPredict, we evaluated more than 40 machine learning classifiers. The best performers, i.e., Nearest Centroid, SVM, and QDA, were further analyzed in detail (see Figure~\ref{fig:Pict8}). The Nearest Centroid method has a modest ability to separate classes and indicate reliably by SVM and QDA, both around 94\%. Gaussian Naive Bayes and Random Forest also demonstrated good performance, with Gaussian Naive Bayes reaching an accuracy of 91\%. The ROC-AUC values supported these findings. Specifically, the Nearest Centroid model achieved:
\begin{itemize}
    \item AUC = 0.95 for Fertile class
    \item AUC = 1.00 for Sub-Fertile class
    \item AUC = 0.97 for Infertile class
\end{itemize}
These high scores demonstrate strong ability to separate classes and indicate reliable classification.
\subsubsection*{Confusion Matrix and ROC Curve Analysis}
The confusion matrix for the Nearest Centroid classifier (as shown in Figure~\ref{fig:Pict9}) demonstrated excellent class-specific performance:
\begin{itemize}
    \item Perfect classification was achieved in the Fast and Slow groups.
    \item One sample in the Average group was misclassified as Slow.
\end{itemize}

\noindent
Despite these minimal misclassifications, the model achieved high accuracy in distinguishing between classes. The misclassifications, especially between the Average and Slow groups, may arise from overlapping morphological or motility traits in borderline samples. This suggests opportunities to improve the model by expanding the training set or including additional biological features. 
The multiclass ROC curve (see Figure~\ref{fig:Pict10}) further validated the model's discriminative power. All curves significantly surpassed the diagonal baseline, and the AUC values confirmed the model’s strong performance across all fertility categories.
\section{Discussion}\label{sec:discussion}
\subsection{Interpretation of Key Findings}
Our study reaffirms the crucial role of sperm concentration, morphology, and progressive motility in evaluating male fertility, consistent with existing literature. Previous research \cite{westerman2020biomarkers} has shown that higher sperm concentration and improved morphology are positively correlated with enhanced motility, underscoring their importance in clinical fertility evaluations. 
Our correlation analysis provides empirical support for this relationship. Specifically, we found that samples with higher sperm concentration and superior morphological characteristics exhibited significantly better progressive motility. This observation aligns with \cite{agarwal2022sperm}, who emphasized the strong relationship between these parameters and fertility outcomes. Furthermore, our results are consistent with \cite{hook2020methodological}, who argued that sperm morphology directly influences motility and, consequently, fertility success rates.

The correlation matrix and visualizations (see Figures~\ref{fig:Pict5}, \ref{fig:pict4}, and \ref{fig:Pict6}) also revealed an interdependence between sperm traits. Progressive motility showed a moderate to strong positive correlation with both morphology and concentration, reinforcing the hypothesis that high-quality sperm are characterized by 
an alignment of multiple parameters rather than isolated features.
\begin{figure}[!htbp]
\centering
\includegraphics[width=0.5\textwidth]{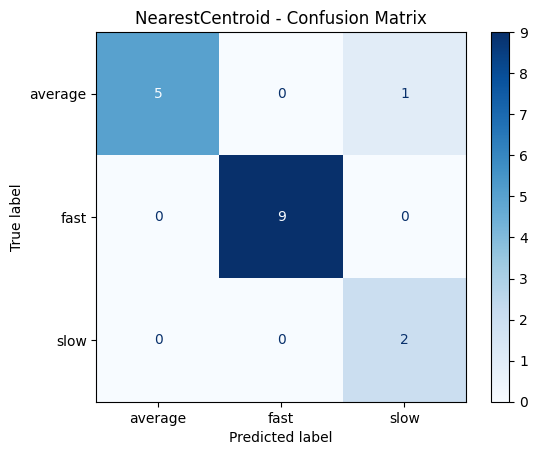}
\caption{%
\textbf{Confusion Matrix for Nearest Centroid Classifier.} 
This confusion matrix shows the Nearest Centroid model's performance across three fertility classes, i.e., Fast, Average, and Slow. The model accurately classified all Fast and Slow samples but misclassified one Average sample as Slow. The color intensity indicates the number of predictions per cell. With an overall accuracy of 94.2\%, the matrix highlights the model's strong ability to differentiate among fertility categories, with minimal errors.
}
\label{fig:Pict9}
\end{figure}
\subsection{Comparative Performance of Classifiers}
In evaluating predictive performance, the Nearest Centroid classifier achieved the highest cross-validated accuracy at 94.2\%, outperforming other models tested using LazyPredict. This finding supports the conclusions of  \cite{hicks2019machine}, who highlighted the effectiveness of centroid-based models in biological classification tasks, especially when data distributions are both compact and distinct.
Although it was slightly less accurate, the Gaussian Naive Bayes model still achieved 91\% accuracy. This result aligns with \cite{wood2019private}, which showed that Naive Bayes performs well in medical and biological domains where input features are only moderately correlated. Given its computational simplicity and interpretability, it remains a valuable option, particularly in clinical decision-support contexts with limited resources.

\begin{figure}[htbp]
\centering
\includegraphics[width=0.5\textwidth]{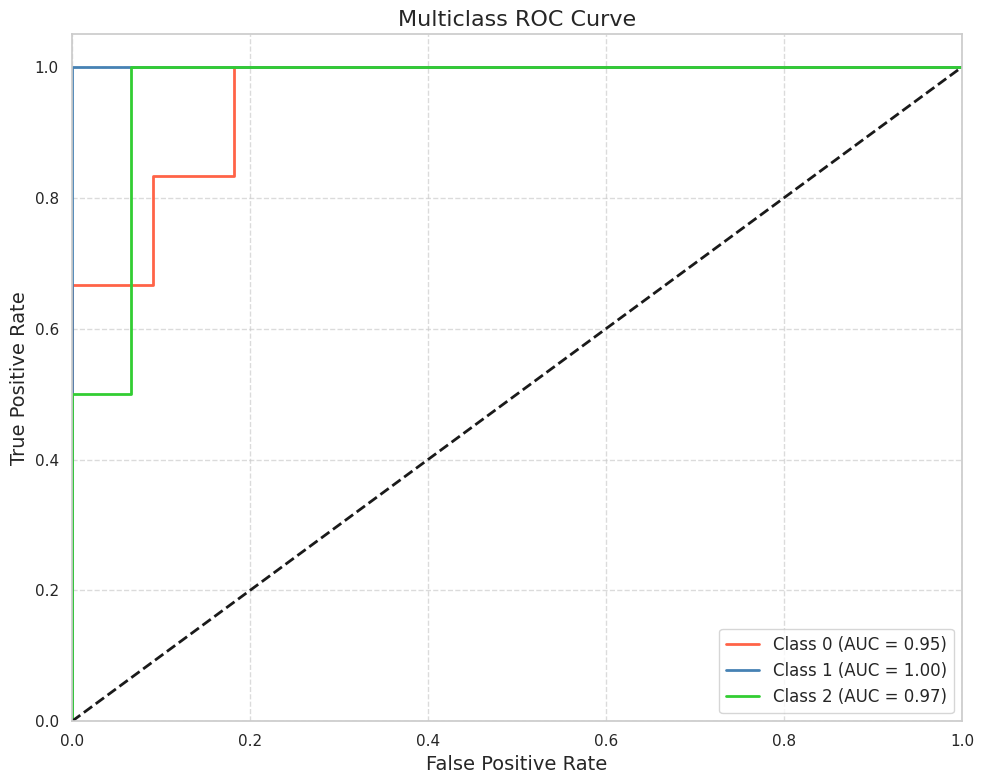}
\caption{%
\textbf{ROC Curves for Fertility Classification Across Three Classes.} 
This figure shows multiclass ROC curves for the Nearest Centroid model. True Positive Rates and False Positive Rates are displayed for each fertility class, i.e., Fertile (AUC = 0.95), Sub-Fertile (AUC = 1.00), and Infertile (AUC = 0.97). The diagonal dashed line indicates the random classification baseline, while the curves highlight the model’s strong predictive ability and excellent class separability.
}
\label{fig:Pict10}
\end{figure}

It is worth noting that all of the top models exhibited strong discriminative power, as indicated by high ROC-AUC scores (see Figure~\ref{fig:Pict10}). The Nearest Centroid model achieved AUC values of 0.95, 1.00, and 0.97 for the Fertile, Sub-Fertile, and Infertile classes, respectively, exceeding the performance thresholds typically expected in fertility diagnostics.

\subsection{Limitations and Areas for Improvement}
While the models demonstrated high overall accuracy, some limitations were identified. The confusion matrix (as shown in Figure~\ref{fig:Pict9}) revealed minor misclassifications, particularly between the Average and Slow categories. 
This overlap may be attributed to morphological and concentration similarities in borderline cases, an observation also noted in \cite{you2021machine}. Although the Nearest Centroid model performed well overall, its sensitivity for false negatives in these adjacent categories needs improvement.
As highlighted by \cite{idowu2015prediction}, minimizing false negatives is critical in clinical contexts. Misclassifying an infertile patient as sub-fertile may lead to inappropriate treatment recommendations. To address this challenge, it may be beneficial to enrich the dataset with a broader range of borderline cases, fine-tune feature selection, or integrate additional biomarkers, such as sperm DNA fragmentation or hormonal profiles.
Moreover, despite the assumption of class separability, fertility data are often influenced by complex, nonlinear interactions and latent variables that are not captured by conventional semen analysis. Future research work could explore the application of ensemble methods or neural networks capable of modeling such nonlinearities, as well as semi-supervised learning techniques to leverage unlabelled clinical data.

Finally, our findings confirm that the Area Under the Curve remains a reliable metric for assessing model performance in medical contexts, as highlighted by \cite{christodoulou2019systematic}. The high AUC scores reported in this study validate the model’s ability to accurately differentiate between fertility categories, reinforcing the potential for machine learning to enhance sperm quality evaluation and inform clinical decision-making.
\section{Conclusion}\label{sec:conclusion}
This study aimed to address a significant challenge in reproductive medicine, i.e., the subjective and often inconsistent evaluation of male fertility through traditional semen analysis. By leveraging machine learning models to classify sperm motility based on various seminal parameters, we demonstrated that automated, data-driven methods can provide more objective, reproducible, and clinically meaningful insights into male reproductive potential.

Our findings showed that models such as Nearest Centroid and Support Vector Machine accurately classified fertility status based on sperm concentration, morphology, and progressive motility, achieving cross-validated accuracies that exceeded 90\%. Visual analyses and correlation matrices further supported these results, revealing strong associations between sperm morphology, motility, and concentration. ROC-AUC evaluations further confirmed the high discriminative power of these models, emphasizing their potential for real-world clinical application.

These findings collectively support the integration of machine learning frameworks into clinical fertility assessment workflows. This integration can help reduce human error, improve diagnostic precision, and provide evidence-based guidance to both patients and clinicians.  Furthermore, expanding these models with additional biological markers, such as hormonal profiles and molecular-genetic indicators, could enhance prediction accuracy and allow for more personalized fertility treatment planning.
More broadly, this research highlights the transformative role of AI in reproductive health, not only in diagnosing male infertility but also in shaping the future of precision medicine.
In short, empowering clinicians with intelligent, data-driven tools may redefine how fertility is understood, diagnosed, and managed, ushering in a new era of reproductive care focused on predictive accuracy and patient-centered outcomes.

\bibliography{references}

\end{document}